\documentclass[preprint,12pt]{elsarticle}

\usepackage{amssymb}
\usepackage{amsmath}

\usepackage{makecell}
\usepackage{multirow}
\usepackage{float}
\usepackage{booktabs}
\usepackage{array} 
\usepackage{booktabs} 
\usepackage{capt-of} 
\usepackage{caption}
\usepackage{tabularx} 

\usepackage[labelfont=bf]{caption}  
\journal{Engineering Applications of Artificial Intelligence}

\begin{document}

\begin{frontmatter}



\title{A multi-task deep learning approach for lane-level pavement performance prediction with segment-level data}


\author{Bo Wang} 
\author{Wenbo Zhang \corref{cor1}}
\author{Yunpeng Li} 
\cortext[cor1]{Corresponding author}
\affiliation{organization={School of Transportation},
            addressline={Southeast University}, 
            city={Nanjing},
            postcode={210000}, 
            state={Jiangsu},
            country={China}}

\begin{abstract}
There is still a lack of elaborate performance analysis at the lane-level due to costly data collection and difficulty in prediction modeling. Therefore, this study developed a multi-task deep learning approach to predict the lane-level pavement performance with a large amount of historical segment-level performance data. In specific, the prediction framework firstly employed an Long Short-Term Memory (LSTM) layer to capture the segment-level pavement deterioration pattern. Then multiple task-specific LSTM layers were designed based on number of lanes to capture lane-level differences in pavement performance. Finally, the multiple task-specific LSTM outputs were concatenated with auxiliary features for further training and the lane-level predictions were obtained after fully connected layer. The aforementioned prediction framework was validated with a real case in China. The proposed prediction framework also outperforms other ensemble learning and shallow machine learning methods in almost every lane.
\end{abstract}

\begin{graphicalabstract}
\begin{figure}[htp]
\centering
\includegraphics[width=1\linewidth]{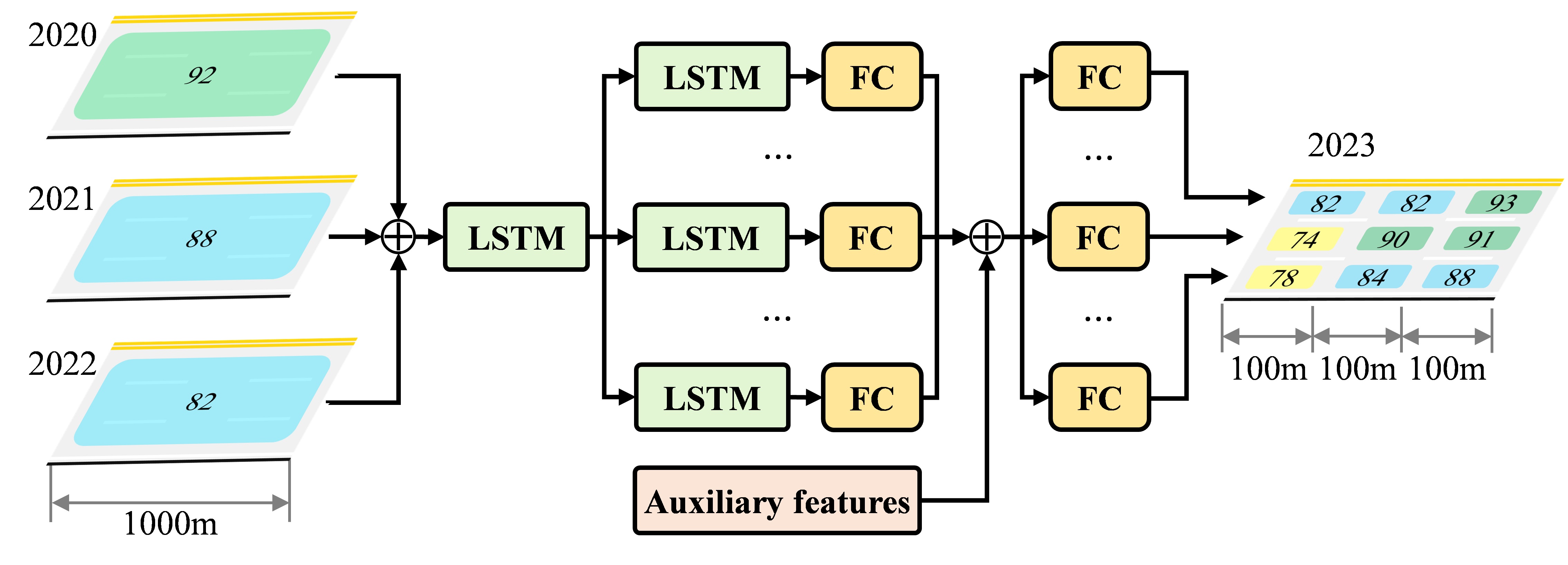}
\end{figure}
\end{graphicalabstract}

\begin{highlights}
\item Propose a deep learning approach for lane-level pavement performance prediction.
\item Introduce LSTM module to investigate temporal relations hidden in historical records.
\item Develop lane-level prediction module using multi-task learning approach.  
\item Concatenate the deep learning module with a set of auxiliary roadway features. 
\item Obtain good model performance under three regular roadway scenarios.
\end{highlights}


\end{frontmatter}



\section{Introduction}
\label{sec1}
The most widely used pavement performance measurements include pavement condition index (PCI), pavement quality index (PQI), and riding quality index (RQI). These measurements in a lane with shorter length can assist in identifying small but severely damaged road sections that are detrimental to traffic. The performance measurement prediction can further provide refined performance trends in future years for each lane. Utilizing the predicted lane-level performance values can further execute the preventive maintenance in a precisely and cost-effective way. Thus, lane-level performance predictions are significantly more useful for pavement maintenance than segment-level ones that merely provide one rough value for all lanes in a segment of 1 km. However, it is impossible to collect a huge amount of lane-level data in reality, despite the fact that many automated pavement detection methods have been proposed\textsuperscript{\cite{-40, -39}}, and it is also difficult to accurately predict lane-level performance in the next few years. To address such difficulty, this study is dedicated to developing a lane-level pavement performance prediction framework with a large amount of segment-level data. 

The different deterioration patterns across lanes are obviously one challenge for developing such unified lane-level prediction framework effective for all lanes. As shown in Figure 1, the difference in performance values across different lanes can range from -60 to 60. The prediction modeling should learn such variations in one unified modeling framework. On the other hand, due to the limited lane-level data but large amount of segment-level data, the data-driven deep learning can be more easier to capture the common deterioration pattern shared by all lanes. It is possible of incorporating the segment-level deterioration pattern into the lane-level prediction to further enhance prediction accuracy. 

\begin{figure}[ht]
\centering
\includegraphics[width=1\linewidth]{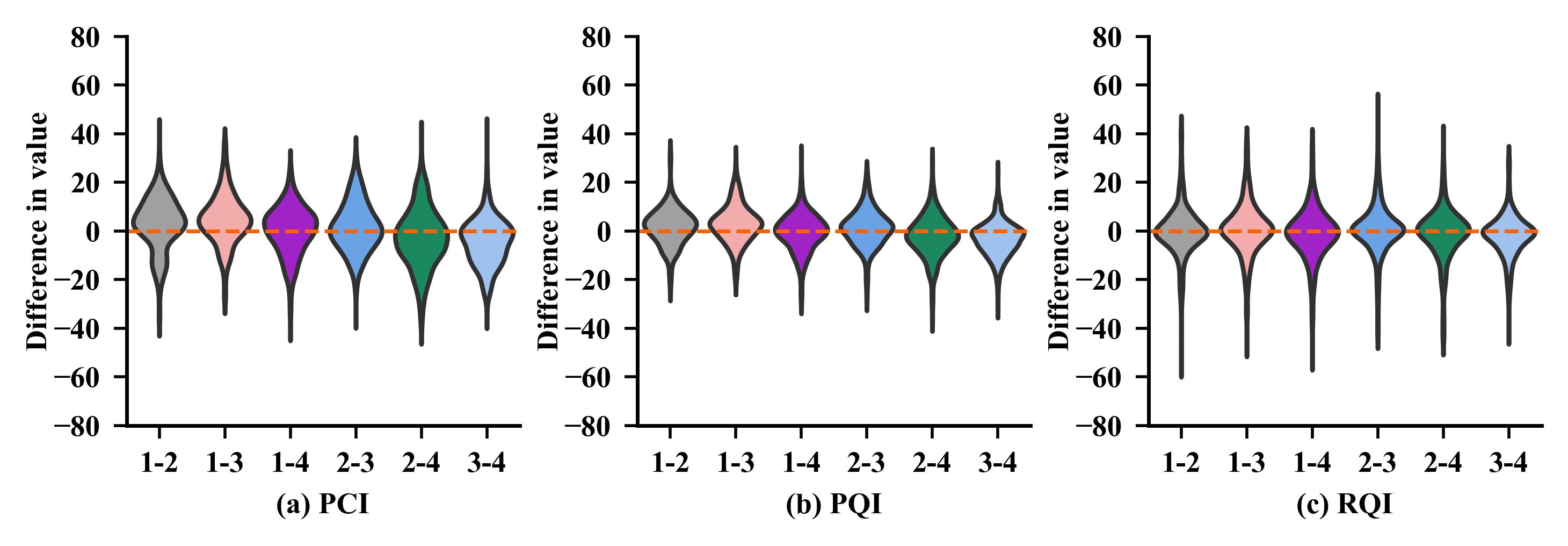}
\caption{The distribution of the difference in PCI, PQI, and RQI between two lanes \\ Note: The ‘1-2’ label on x-axis means the performance difference between lane 1 and lane 2. Lane 1 is the innermost lane. This is applicable to other labels.}\label{fig1}
\end{figure}

The state-of-the-art studies have already proposed models for segment-level pavement performance prediction while existing fewer lane-level prediction modeling frameworks. Few of them also developed machine learning-based methods, such as artificial neural networks\textsuperscript{\cite{14}} and random forest\textsuperscript{\cite{-29}}, which yield good prediction accuracy. However, these models are not discussed at lane-level and cannot capture the differences across lanes. 

To address the difficulties during lane-level performance prediction, a multi-task deep learning approach is proposed for lane-level pavement performance prediction based on segment-level data to deal with those challenges. The model employs an LSTM-based shared layer to capture the common and representative changing pattern information from pavement performance time series. Based on the shared information, the task-specific heads are used to extract the details and changes related to a specific lane and generate predictions for the lane. Subsequently, the predictions are horizontally concatenated with auxiliary features, and the features distinguish roads with different decay patterns. At last, the task-specific output layer will handle the concatenation results to get the final lane-level predictions. To verify the proposed model, the real case was employed to check prediction accuracy of PCI, PQI, and RQI on three scenarios including one-way 2-lane, 3-lane, and 4-lane roads.

The remainder of this paper is organized as follows: In section 2, the related work on pavement performance prediction is summarized. In section 3, the structure of the proposed model is introduced detailed. Then, in section 4, the model is validated through comparative and ablation experiments based on a dataset that has been independently acquired. Finally, conclusions and guidelines for future work were presented.

\section{Related work}
\label{sec2}
The literature on pavement performance prediction models can be categorized into two typical groups of probabilistic reasoning models and shallow machine learning models. The group of probabilistic reasoning models primarily relies on specific mathematical equations or statistical laws. The group of shallow machine learning models contains typical data-driven models, rather than deriving from theoretical assumptions or prior knowledge.

\subsection{Probabilistic reasoning models}
Probabilistic reasoning models predict pavement performance considering uncertainty and probability distribution of influencing factors. Specifically, the models can be further divided into empirical, mechanistic-empirical, and probabilistic models. Empirical models are based on traditional regression methods whose parameter are estimated with one dataset of pavement performance and other factors. Cao selected service life and cumulative number of axle loads as independent variables and applied cosine deterioration equations to predict PCI and rutting depth index of roads\textsuperscript{\cite{2}}. Wu examined the better performance of the S-shaped model on pavement performance prediction, comparing with the polynomial and exponential models\textsuperscript{\cite{1}}. Mechanistic-empirical models are usually based on mechanical principles and explore the relationship between pavement performance and pavement responses (i.e. stresses, strains, and deflections). Sidess combined a mechanistic-empirical approach with a regressive empirical approach to predict pavement performance \textsuperscript{\cite{9}}. Probabilistic models consider significant factors in pavement performance deterioration and describe the possible distribution or changing trend of pavement performance by statistical methods, such as Markov models and Monte Carlo simulations. Wang used the gray theory to predict pavement smoothness and used the fuzzy regression method to determine the coefficients for the gray prediction model\textsuperscript{\cite{20}}. Dong predicted asphalt pavement performance utilizing Markov models based on insufficient inspection data\textsuperscript{\cite{4}}. Wang put forward a Grey–Markov combination model to predict PCI\textsuperscript{\cite{5}}. Since the pavement deterioration rate may not be constant in the long term, Abaza developed a non-homogeneous Markov chain with different transition matrices\textsuperscript{\cite{6}}. Besides, Wasiq proposed a hybrid approach combining homogeneous and non-homogeneous Markov Chain to predict the PCI\textsuperscript{\cite{7}}. 

Probabilistic reasoning models perform well in particular with limited observation data, since such models assume underlying pavement deterioration pattern with classical distributions or equations, even without any proofs. The model performance mainly depends on very reasonable assumptions or statistically validated hypothesis. However, the pavement deterioration may be influenced by various factors including location-specific, lane-specific, time-specific, pavement design, and traffic. It is challenging in finding an universal pavement deterioration pattern applicable anywhere and anytime. Thus, the probabilistic reasoning models require a great deal of time and effort for deriving correct assumptions on pavement deterioration while predicting lane-level performance in a large-scale road network.

\subsection{Machine learning models}
Machine learning models are new trends in pavement performance prediction, typically including traditional machine learning algorithms and artificial neural networks (ANN). Such models are trained only with observation datasets, generally not introducing any assumptions or predefined equations/distributions. The trained modeling structure, as well as corresponding parameters, reveals the nonlinear relationships underlying the pavement deterioration. Regarding the traditional machine learning algorithms, Wang developed a hybrid grey relation analysis and support vector machine regression model to predict pavement performance\textsuperscript{\cite{11}}. Hu proposed a random forest-based prediction model and further enhanced model perforamnce with a grid search algorithm\textsuperscript{\cite{20}}. Ali investigated the combined effect of pavement distress on flexible pavement performance in two climate regions (wet freeze and wet no freeze) separated using multiple linear regression and artificial neural network\textsuperscript{\cite{14}}. Zhang and Kaloop both employed the Gaussian process regression model for predicting IRI. The distinction lies in that Zhang utilized a genetic algorithm to determine the optimal time series length\textsuperscript{\cite{34}}, whereas Kaloop adopted regression modeling to select appropriate feature inputs\textsuperscript{\cite{35}}. In addition, the ensemble learning methods are also used to predict pavement performance due to their good performance in complicated industry problems, such as XGBoost\textsuperscript{\cite{19}} and ThunderGBM\textsuperscript{\cite{29}}.

Regarding the deep learning based methods, the BPNN modeling framework was further improved with a differential evolution particle swarm optimization by Tao, while predicting PCI\textsuperscript{\cite{13}}. Cai proposed a causal-temporal graph convolution network, combining the graph convolution networks and LSTMs to simultaneously capture causal features and temporal features to predict pavement performance\textsuperscript{\cite{15}}. Bukharin proposed a two-stage model combining LSTM and ANN. The LSTM is used to learn the pavement deterioration pattern from time-series data, and the ANN further incorporates influencing factors for deterioration pattern\textsuperscript{\cite{16}}. Guo proposed a weighted multi-output neural network to predict IRI, faulting, longitudinal crack, and transverse crack in one unified framework\textsuperscript{\cite{17}}. Li proposed a model combining the BPNN and the LSTM to predict the highway pavement performance with a high precision of 100 m, based on ten-year traffic load data, climatic history, and maintenance records data\textsuperscript{\cite{18}}. Zhang developed a graph deep learning framework for network-based traffic prediction, and the frame models the dynamics of the traffic flow on a road network as a Markov chain on a directed graph. Still, it can not predict lane-level traffic volume which is important, especially in intersections.\textsuperscript{\cite{31}}. Yao used a deep reinforcement learning model to learn better maintenance strategies that maximize long-term cost-effectiveness in maintenance decision-making. In particular, each lane can have different treatments, and the long-term maintenance cost-effectiveness of the entire road is treated as the optimization goal. However, the lane-level data the model needs limits the wide use of the model\textsuperscript{\cite{32}}. 

The machine learning models are good at investigating and capturing hidden relationships in observation data, always yielding better performance once having enough big data. These models have been validated in the task of segment-level pavement performance prediction. However, the literature lacks examples of similar modeling structures for lane-level pavement performance prediction according to the authors' investigation, in particular one unified prediction model resulting in pavement performance of multiple lanes. The task of lane-level pavement performance prediction is challenging due to much more complicated correlations and differences among lanes, compared to the task of segment-level prediction.

\section{Method}
\subsection{Problem definition}
This study is dedicated to proposing a lane-level pavement performance prediction model with both historical segment-level pavement performance data and auxiliary features influencing pavement deterioration.
\begin{equation} 
\mathbf{\hat{y}^{t+1}_{n,~j}} = f(\mathbf{y^{t-k:t}_{m}},~\mathbf{X_m}),~n=1, 2, ..., N. 
\end{equation} 
where,\\
\indent$\mathbf{\hat{y}^{t+1}_{n,~j}}$ is the predicted pavement performance, like PCI, PQI, or RQI,  on spatial prediction unit $j$ of lane $n$ at time step $t+1$; \\
\indent$n$ is the lane number, ranged from 1 to $N$; \\
\indent$N$ is the total number of lanes in one-way, usually 1, 2, or more;\\
\indent$j$ is the spatial prediction unit number generally with a length of $100m$;\\
\indent$t$ is the time step, generally representing the year;\\
\indent$\mathbf{y^{t-k:t}_{m}}$ is the time series of pavement performance observed on spatial unit $m$ (i.e. one segment) from time step $t-k$ to time step $t$;\\
\indent$\mathbf{X_{m}}$ is the auxiliary features measured on spatial unit $m$, in this study auxiliary features are time-invariant;\\
\indent$m$ is the spatial measurement unit number, each unit is one road segment with a length of $1000m$, including multiple spatial prediction units $j$;\\ 
\indent$f(\cdot)$ is the trained modeling structure.

\subsection{Model overview}
The section is focus on introducing the structure of the proposed multi-task deep learning model for lane-level pavement performance prediction, as shown in Figure 2. First, input features are processed with one segment-level deep learning module (a) that is based on LSTM, extracting general deterioration patterns at segment level from historical pavement performance data. Second, one task-specific deep learning module (b) yielding multiple heads for each lane-level prediction task and further learning underlying patterns with LSTM and fully connected neural network layers. In final, one concatenation module (c) horizontally mixes deep learning output with auxiliary features, resulting in final lane-level predictions by multiple fully connected neural networks. Regarding the model training, the mean square error (MSE) and their total sum are used to quantify the loss of each task and all tasks, respectively. 

\begin{figure}[htp]
\centering
\includegraphics[width=0.8\linewidth]{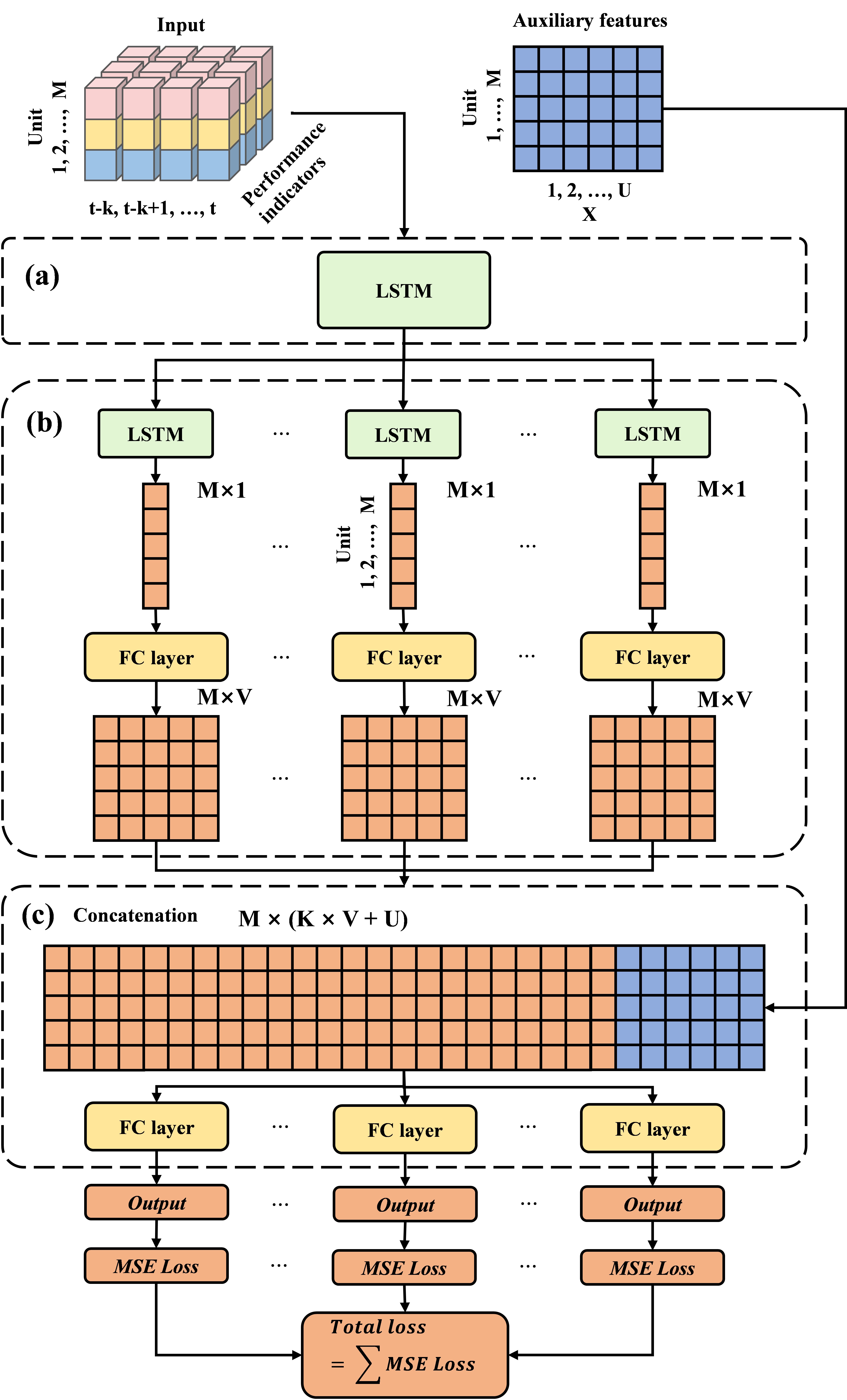}
\caption{The overall lane-level prediction modeling structure}\label{fig2}
\end{figure}

\subsubsection{LSTM-based shared layer}
The LSTM-based shared layer is initially used to handle the pavement performance time series at segment level. It is responsible for discovering and extracting common deterioration patterns of the whole road segment. LSTM is a recurrent neural network architecture designed to handle the vanishing and exploding gradient problems often encountered in traditional recurrent neural networks when dealing with long data sequences. The LSTM has a more complex internal structure than standard recurrent neural network cells. It consists of three main gates including the input gate, the forget gate, and the output gate. The input gate decides how much new information to let into the cell state. The forget gate determines which parts of the previous cell state to discard. The output gate controls how much of the cell state is output as the hidden state. The aforementioned gate mechanism allows LSTM to selectively remember or forget information, making it effective in long-term prediction tasks.

(1) Forget gate\\
\indent The forget gate controls what information is removed from the memory cell. The gate takes two inputs of segment-level pavement performance $y^{tt}$ and hidden state at the previous time step $tt-1$. The both inputs are multiplied with one weight matrix $W_{g}$, which further adds one bias $b_{g}$, as in equation 2. The result is passed through an activation function that yields an output between 0 and 1. In specific, if the output is 0, the information is forgotten. In contrast, if the output is 1, the information is retained for further computation in future time steps.
\begin{equation}
g_{tt} = \sigma(W_{g}[h_{tt-1}, y^{tt}] + b_{g})
\end{equation}
where, \\
\indent$y^{tt}$ is set of pavement performance on road segment $1$ to $M$ at time step $tt$, $\left[y^{tt}_{1}, y^{tt}_{2}, \ldots, y^{tt}_{M} \right]$;\\
\indent$g_{tt}$ is the forget gate information indicator at time step $tt$;\\
\indent$\sigma(\cdot)$ is the sigmoid activation function;\\
\indent$W_g$ represents the weight matrix used in the forget gate;\\
\indent$h_{tt-1}$ is the hidden state at the previous time step $tt-1$, integrating information from the first time step to the time step $tt-1$;\\
\indent$b_{g}$ is the bias in the forget gate.\\

(2) Input gate\\
\indent The input gate controls what information is added to each memory cell. On one hand, the information indicator $i_{tt}$ is initially computed with weighted inputs and sigmoid function, similar as computation in forget gate. On the other hand, the initial cell state $\stackrel{\sim}{c_{tt}}$ is also computed with weighted inputs, but is activated with the tanh function, shown in equation 4. Such activation function will result in an output ranging from -1 to 1. The cell state $c_{tt}$ is finally obtained with equation 5, combining together cell state at previous time step, the forget gate output at current time step, the new information in input gate, and the initial cell state.  
\begin{equation} 
i_{tt} = \sigma(W_{i}[h_{tt-1}, y^{tt}] + b_{i})
\end{equation} 
\begin{equation} 
\stackrel{\sim}{c_{tt}} = tanh(W_{c}[h_{tt-1}, y^{tt}] + b_{c})
\end{equation} 
\begin{equation} 
c_{tt} = g_{tt} \odot c_{tt-1} + i_{tt} \odot \stackrel{\sim}{c_{tt}}
\end{equation} 
where,\\
\indent $i_{tt}$ is information indicator in the input gate at time step $tt$;\\
\indent $W_i$ is the weight matrix in the input gate;\\
\indent $b_{i}$ is the bias in the input gate;\\
\indent $\stackrel{\sim}{c_{tt}}$ is the initial cell state at time step $tt$;\\
\indent $\tanh(\cdot)$ is the tanh activation function;\\
\indent $W_c$ is the weight matrix in the memory cell;\\
\indent $b_c$ is the bias in the memory cell;\\
\indent $c_{tt}$ is the cell state at time step $tt$;\\
\indent $c_{tt-1}$ is the cell state at time step $tt-1$;\\
\indent $\odot$ is the Hadamard product denoting an element-wise product.\\

\indent (3) Output gate:\\
\indent The output gate controls what information is output from the memory cell. First, the output information indicator $o_{tt}$ is generated through activating weighted inputs as other information indicators. Then, the hidden state $h_{tt}$ will be updated by the output information indicator and the tanh activation of cell state, as follows.
\begin{equation} 
o_{tt} = \sigma(W_{o}[h_{tt-1}, y^{tt}] + b_{o})
\end{equation} 
\begin{equation} 
h_{tt} = o_{tt} \odot tanh(c_{tt})
\end{equation} 
\noindent where,\\
\indent$o_{tt}$ is the output information indicator in the output gate at time step $tt$;\\
\indent$W_o$ represents the weight matrix in the output gate;\\
\indent$b_{o}$ is the bias in the output gate;\\
\indent $h_{tt}$ is the hidden state at time step $tt$.

\subsubsection{Multiple task-specific heads}
The module (b) in Figure 2 provides multiple task-specific heads, designed to learn unique pavement performance deterioration patterns for each lane. Each head comprises an LSTM unit and a fully connected neural network layer. The LSTM component is important in capturing pavement performance of each lane by taking hidden output and cell states from the LSTM-based shared layer. The following fully connected neural network layer maps the LSTM predictions to a tailored shape.
\begin{equation} 
\mathbf{S_{n}} = \mathbf{s_{n}} * \mathbf{W_{x}}, \mathbf{S_{n}\in\mathbb{R}^{M \times V}}, \mathbf{s_{n}\in\mathbb{R}^{M \times 1}}
\end{equation}
where,\\
\indent $\mathbf{{S}_{n}}$ is the fully connected layer's output at lane $n$; \\
\indent $\mathbf{{s}_{n}}$ is the LSTM output at lane $n$; \\
\indent $\mathbf{W_{s}}$ is the weight matrix of the fully connected neural network layer; \\
\indent $V$ is the shape size of one output dimension in the fully connected neural network layer.\\


In the fully connected neural network layer, the LeakyReLU as activation function is used, as in equation 9. Since such activation function has a non-zero gradient for negative inputs, which makes the gradients propagation more effectively in the deep neural network.
\begin{equation}
\hfill
    LReLU(Val) =
    \begin{cases}
    Val, & \text{if } Val \geq 0 \\
    \alpha Val, & \text{if } Val < 0
    \end{cases}
\hfill
\end{equation}
where,\\
\indent $\alpha$ is a small constant, often ranging from 0.01 and 0.1.\\

\subsubsection{Multiple task-specific output layers}
The module (c) in Figure 2 first horizontally concatenate the output of module (b) and auxiliary features. Then the fully connected neural network layer transforms the concatenation tensor into lane-level predictions.
\begin{equation}
\mathbf{X} = \left[\mathbf{X_{1}}, \mathbf{X_{2}}, \ldots, \mathbf{X_{m}}, \ldots, \mathbf{X_{M}} \right],~ \mathbf{X_{m}} \in \mathbf{R}^{U \times 1},~\mathbf{X} \in \mathbb{R}^{M \times U}
\end{equation}
\begin{equation}
\mathbf{S} = \left[\mathbf{S_{1}},\ldots, \mathbf{S_{n}}, \ldots, \mathbf{S_{N}}, \mathbf{X} \right],~\mathbf{S_{n}} \in \mathbf{R}^{M \times V}
\end{equation}
\begin{equation}
\mathbf{\widehat{Y}^{t+1}_{n}} = \mathbf{S} \times \mathbf{W_{n}},~\mathbf{\widehat{Y}^{t+1}_{n}} \in \mathbf{R}^{M \times 1}
\end{equation}

\noindent where,\\
\indent $\mathbf{X}$ is the matrix of auxiliary features measured at all $M$ segments; \\
\indent $U$ is the number of auxiliary features; \\
\indent $\mathbf{S}$ is the concatenation result of module (b) output and auxiliary features; \\
\indent $\mathbf{\widehat{Y}^{t+1}_{n}}$ is the final pavement performance predictions of lane $n$ at time step $t+1$;\\ 
\indent $\mathbf{W_{n}}$ is the weight matrix in the fully connected neural network layer for lane $n$.

\subsubsection{Total loss calculation}
In the proposed multi-task learning-based model, the parameter estimation relies on the training loss of not only each lane-specific prediction, but also the whole sample prediction. In this study, mean squared error (MSE) is used as the loss function for each lane-specific prediction, to derive a total summation of all MSEs as the total loss.
\begin{equation}
L = L_1 + L_2 + \dots + L_n + \dots + L_N
\end{equation}
\begin{equation}
L_{n} = \frac{1}{M} \sum_{j = 1}^{M} (\boldsymbol{y}^{t+1}_{n,~j} - \widehat{\boldsymbol{y}}^{t+1}_{n,~j})^2
\end{equation}
where,\\
\indent $L$ is the overall pavement prediction MSE for model training; \\
\indent $L_n$ is the pavement prediction MSE of lane n; \\
\indent $\boldsymbol{y}^{t+1}_{n,~j}$ is the actual pavement performance on spatial prediction unit $j$ of lane $n$ at time step $t+1$.

\section{Results}
\subsection{Dataset and preprocessing}
One real pavement performance measurement dataset is collected in three cities of Zhengzhou, Jiaozuo, and Luoyang in Henan Province, China, as shown in Figure 3. The pavement performance is mainly evaluated with three common indicators including PCI, PQI, and RQI, ranging from the year of 2020 to the year of 2023. The data in the first three years is measured on each road segment with a length of around $1000m$ or 0.6 mile. The data in the last year is measured on each lane, dividing into units with a length of $100m$ or 0.06 mile. Meanwhile, a subset of auxiliary features has been collated, including road names, direction(upstream or downstream), road classification, city and town information, pavement surface material, traffic volume, and speed limit. All features, as well as their statistics, are summarized in TABLE 1. 

\begin{figure}[htp]
\centering
\includegraphics[width=1\textwidth]{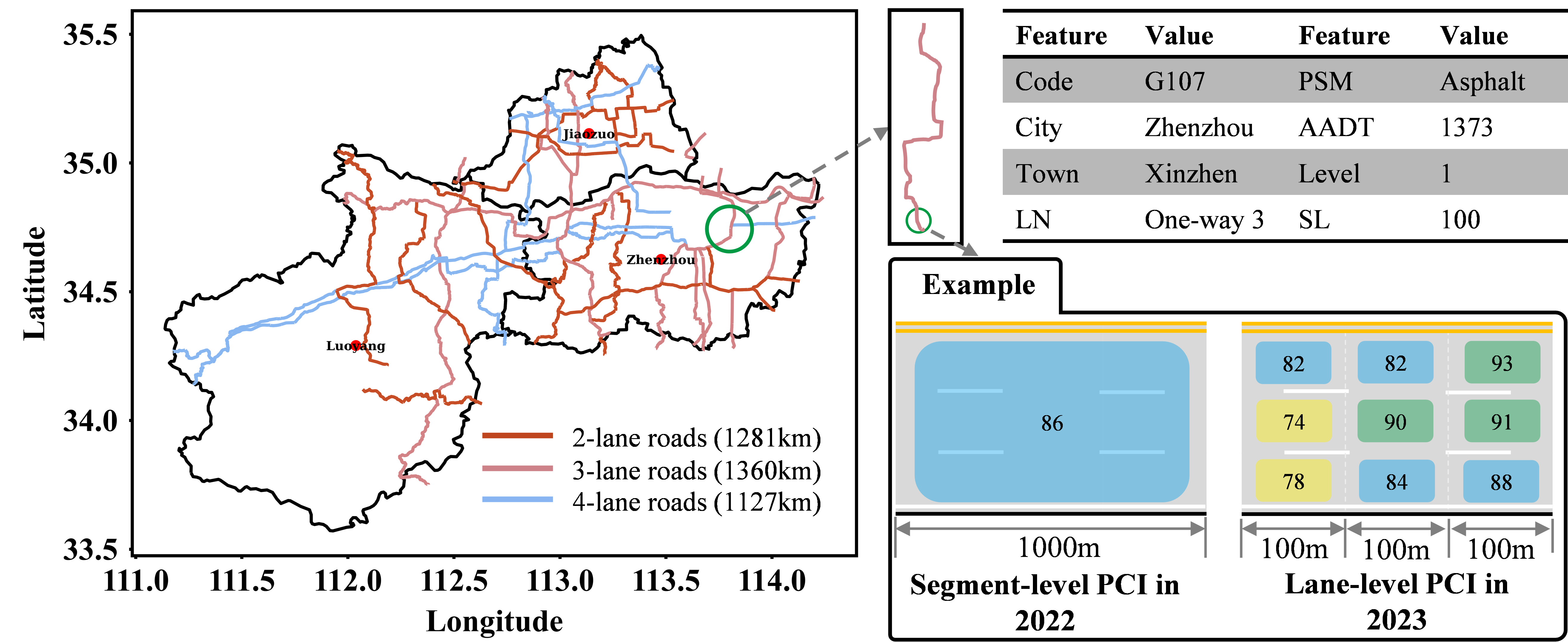}
\caption{The description of the study case in Henan Province, China}\label{fig3}
\end{figure}

\begin{table}
\begin{center}
\caption{The features collected in the study case}
\resizebox{1.0\linewidth}{!}{
\begin{tabular}{l l l l l}
\hline
No. & Feature name & Description & Type & Ranges\\
\hline
    1 & PCI & Pavement Condition Index & Continuous & From 0 to 100\\
    2 & PQI & Pavement Quality Index & Continuous & From 0 to 100 \\
    3 & RQI & Riding Quality Index & Continuous & From 0 to 100 \\
    4 & Code & Road name & Discrete & G107, S323, etc. \\
    5 & Dir. & \makecell[tl]{Upstream or downstream \\ direction of the road} & Discrete & Upstream, downstream\\
    6 & Level & \makecell[tl]{The classification of road} & Discrete & From 1 to 4 \\
    7 & City & \makecell[tl]{The city where the \\ road segment is located} & Discrete & \makecell[tl]{Zhenzhou, Luoyang, \\Jiaozuo city} \\
    8 & Town & \makecell[tl]{The town where the \\ road segment is located} & Discrete & \makecell[tl]{Jinshui district, Erqi \\ district, etc.}\\
    9 & SL & Speed limit of the road & Continuous & From 60 to 120  \\
    10 & AADT & Average Annual Daily Traffic & Continuous & From 50 to 3000\\
    11 & PSM & pavement surface material & Discrete & Asphalt, concrete \\
    12 & LN & \makecell[tl]{Lane number, and the \\ innermost is lane 1} & Discrete & From 1 to 4 \\
\hline
\end{tabular}
}
\end{center}
Note: discrete features will be encoded using one-hot encoding.
\end{table}

Due to the data quality and maintenance, data cleaning was applied to the real dataset. First, deleting the road segments with the implementation of maintenance during data collection periods. Since the maintenance will significantly change the pavement performance in a short time. Second, measurement errors may result in fluctuations, violating year-by-year degradation pattern. The road segments with such fluctuations are also excluded from samples. In final, the whole dataset is divided into three distinct datasets based on three different road scenarios of one-way 2-lane, 3-lane, and 4-lane. The statistics of all three datasets are summarized in TABLE 2. Each dataset will be partitioned into training and testing set by a ratio of 7:3. The auxiliary features will undergo Z-score normalization before model training and testing. 

\begin{table}
\caption{The number of data samples in three road scenarios}\label{tab2}
\begin{tabular*}{\textwidth}{@{\extracolsep{\fill}}l l l l}
\toprule
Dataset & PCI & PQI & RQI \\\hline
2-lane dataset & 524 & 536 & 516\\
3-lane dataset & 705 & 792 & 678\\
4-lane dataset & 716 & 812 & 848\\\hline
\end{tabular*}
\end{table}

\subsection{Model performance evaluation}
Mean absolute percentage error(MAPE) is employed to assess the performance of the proposed model. MAPE reflects the average relative deviation of the prediction values from the actual values. Thus the smaller their values, the better the model's performance. The MAPE of lane $n$ prediction task is the average absolute relative deviation of the prediction performance values from the actual performance values on the lane $n$. The overall MAPE is the average of MAPE on all lanes. 

\begin{equation}
MAPE_{n} = \frac{1}{M} \sum_{j=1}^{M}\left| \frac{y^{t+1}_{n,j} - {\hat{y}^{t+1}_{n,j}}}{y^{t+1}_{n,j}} \right| \times 100\%
\end{equation}
\begin{equation}
    MAPE = \frac{1}{N} \sum_{n=1}^{N} MAPE_{n}
\end{equation}
where,\\
\indent$MAPE_{n}$ is the MAPE on lane $n$; \\
\indent$\hat{y}^{t+1}_{n,j}$ is the predicted pavement performance value of unit $j$ on lane $n$ at time step $t+1$; \\
\indent$y^{t+1}_{n,j}$ is the actual pavement performance value of unit $j$ on lane $n$ at time step $t+1$;\\
\indent$MAPE$ is the overall MAPE of all lanes.\\

\subsection{Model setting}
The model parameters can be set as follows. The LSTM-based shared layer includes two layers with a hidden layer of dimension 128. In the multiple task-specific heads, the LSTM contains two layers with a hidden layer of dimension 64, and a fully connected layer has two layers with a hidden layer of dimension 64 and an output layer of dimension 32. The activation function employed in the fully connected layers is the LeakyReLU. The number of task-specific heads is variable, aligning with the number of one-way lanes, and usually ranges from two to four. The fully connected layer in the multiple task-specific output layers is size 32 and output dimension 1. The hyperparameters that underpin the proposed model include the learning rate, batch size, and the number of training epochs that are 0.001, 32, and 200, respectively. The output shape of each layer is shown in Table 3.

\begin{table}
\caption{Output shape of typical model components}\label{table 3}
\begin{tabular*}{\textwidth}{@{\extracolsep{\fill}}l l l l}
\toprule
No. & Layer & Output shape \\
\toprule
1 & Input & $ M \times 3 \times 1 $ \\
2 & LSTM based shared layer & $ M \times 3 \times 1 $ \\
3 & LSTM in task-specific head & $ M \times 1 $ \\
4 & Fully connected layer in task-specific head & $ M \times 32 $ \\
5 & Task-specific output layer & $ M \times 1 $ \\
\toprule
\end{tabular*}\\
Note: $m$ is the number of prediction units.
\end{table}

\subsection{Model performance}
The study encompasses an evaluation of four benchmark models, comprising the Long Short-Term Memory network (LSTM), Gated Recurrent Unit (GRU), eXtreme Gradient Boosting Regressor (XGBoost), and Random Forest (RF). Furthermore, the study involves the establishment of two distinct experimental configurations: the 'lane-specific model' and the 'mixed model'. The lane-specific model will train a dedicated model for each lane. The 'mix model' predict pavement performance of all lanes in one unified modeling framework. In all three road scenarios of one-way 2-lane, 3-lane, and 4-lane, the proposed model demonstrates superior performance in terms of MAPE regardless of predicting PCI, PQI, and RQI. The MAPE results of all tests are presented in Table 4.

\begin{table}[ht]
    \caption{The MAPE of proposed model and benchmark models}\label{tab4}
    \resizebox{1.0\linewidth}{!}{
    \begin{tabular}{lllllllllll}
        \toprule 
        & & \multicolumn{3}{c}{2-lane dataset} & \multicolumn{3}{c}{3-lane dataset} & \multicolumn{3}{c}{4-lane dataset}\\
        \cmidrule(lr){3-5} \cmidrule(lr){6-8} \cmidrule(lr){9-11}
        configurations & model & PCI & PQI & RQI & PCI & PQI & RQI & PCI & PQI & RQI \\
        \midrule
        \multirow{4}{*}{\makecell[tl]{Lane \\ specific \\ model}}& LSTM & 12.554 & 15.133 & 10.543 & 6.743 & 8.221 & 7.862 & 8.382 & 7.600 & 7.602 \\
        & GRU & 12.239 & 14.639 & 9.790 & 6.623 & 9.577 & 7.970 & 8.311 & 7.319 & 5.819 \\
        & XGBoost & 11.977 & 14.202 & 9.985 & 6.199 & 8.711 & 7.063 & 9.073 & 7.605 & 5.440 \\
        & RF & 11.707 & 13.973 & 10.057 & 5.946 & 8.419 & 6.698 & 9.952 & 7.076 & 5.693  \\
        
        \multirow{4}{*}{Mix model}& LSTM & 13.525 & 12.495 & 10.520 & 6.357 & 8.017 & 6.262 & 8.685 & 7.514 & 5.704\\
        & GRU & 12.685 & 11.979 & 10.484 & 6.104 & 7.740 & 6.217 & 10.389 & 6.867 & 5.741\\
        & XGBoost & 11.661 & 11.731 & 10.612 & 5.880 & 7.771 & 6.310 & 10.169 & 6.765 & 6.029 \\
        & RF & 11.327 & 11.458 & 10.276 & 5.866 & 7.468 & 5.849 & 10.499 & 6.580 & 5.242  \\

        \midrule
        \multirow{1}{*}{MTL model}& Ours  & \textbf{10.799} & \textbf{10.837} & \textbf{9.291} & \textbf{5.273} & \textbf{6.980} & \textbf{5.195} & \textbf{5.752} & \textbf{5.739} & \textbf{3.317} \\

        \bottomrule
    \end{tabular}}
\end{table}

(1) The proposed model achieves an average of 10.309\%, 5.816\%, and 4.936\% MAPE when predicting PCI, PQI, and RQI on the 2-lane, 3-lane, and 4-lane datasets, respectively. The outcome is related to the quantity of prediction units and tasks. The greater the number of units, the more common and unique pavement decay patterns the model acquires. Furthermore, a greater number of lanes is more conducive to enhancing the process of information sharing.

(2) Compared with the 'lane-specific model' and 'mix model', the proposed model has accomplished an average absolute maximum reduction in MAPE of 2.774\%, 2.699\%, and 3.092\% on the 2-lane, 3-lane, and 4-lane datasets, respectively. Correspondingly, in sequence, the average absolute minimum reduction in MAPE is 0.560\%, 0.556\%, and 1.480\% on the datasets mentioned above. The outcomes show the effectiveness of the model in leveraging related tasks to enhance the overall forecasting accuracy in pavement performance analysis. Besides, the MAPE reduction on the 2-lane and 3-lane datasets is closed, while on the 4-lane dataset, the reduction is almost three times that of the reduction on the 2-lane and 3-lane datasets. It demonstrates that increasing the number of tasks does not necessarily increase the magnitude of MAPE reduction.

(3) The model attains an average MAPE of 7.852\%, 7.275\%, and 5.934\%, respectively, for predicting PQI, PCI, and RQI. Furthermore, as depicted in Figure 1, the performance values of different lanes is different, and the dataset exhibits the maximum average PQI difference of 2.834 between different lanes, succeeded by PCI with a difference of 1.950 and RQI with a difference of 0.946. The MAPE demonstrates a specific positive correlation with the performance disparity between lanes. The positive correlation indicates that the similarity between lanes and the resemblance between tasks can mitigate the model's prediction error.

(4) Figure 4 demonstrates lane-level MAPE results using the proposed model. Moreover, as the number of lanes escalates, the difference in MAPE among lanes narrows. Additionally, the MAPE of lane 1 and lane 2 generally shows a downward trend because the information of the newly added prediction tasks will improve the accuracy of the existing performance prediction tasks.

\begin{figure}[htp]
\centering
\includegraphics{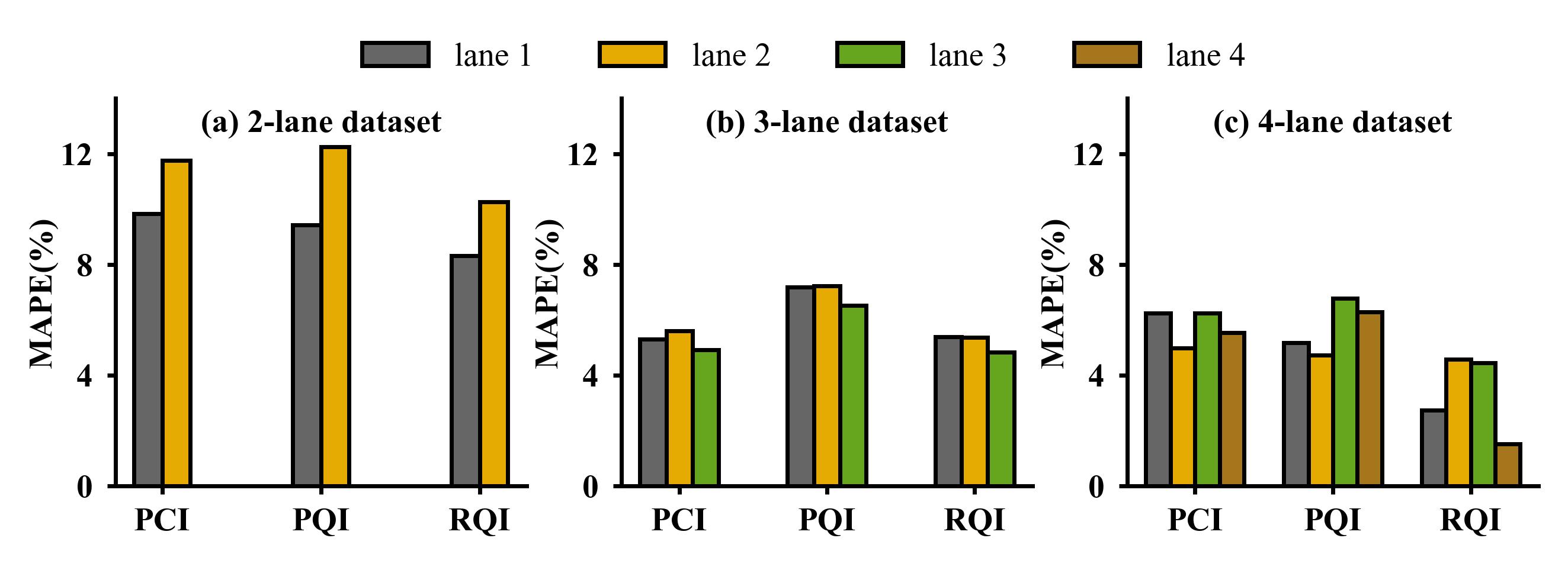}
\caption{The lane-level MAPE of our model}\label{fig4}
\end{figure}

(5) Figures 5, 6, and 7 show lane-level MAPE based on 2-lane, 3-lane, and 4-lane datasets. In most cases, the MAPE of each lane is the lowest using our model compared with the lane-specific and mix models. In the 2-lane dataset, there are 3 times, accounting for 50\%, that the lane-level MAPE is not the lowest. And all are from lane 2. But in 3-lane and 4-lane datasets, the numbers that the lane-level MAPE is not the lowest are 2 and 4, only accounting for 22\% and 33\%. The result means that the multi-task model is suitable for decreasing the MAPE of each lane.

\begin{figure}[htp]
\centering
\includegraphics[width=1\textwidth]{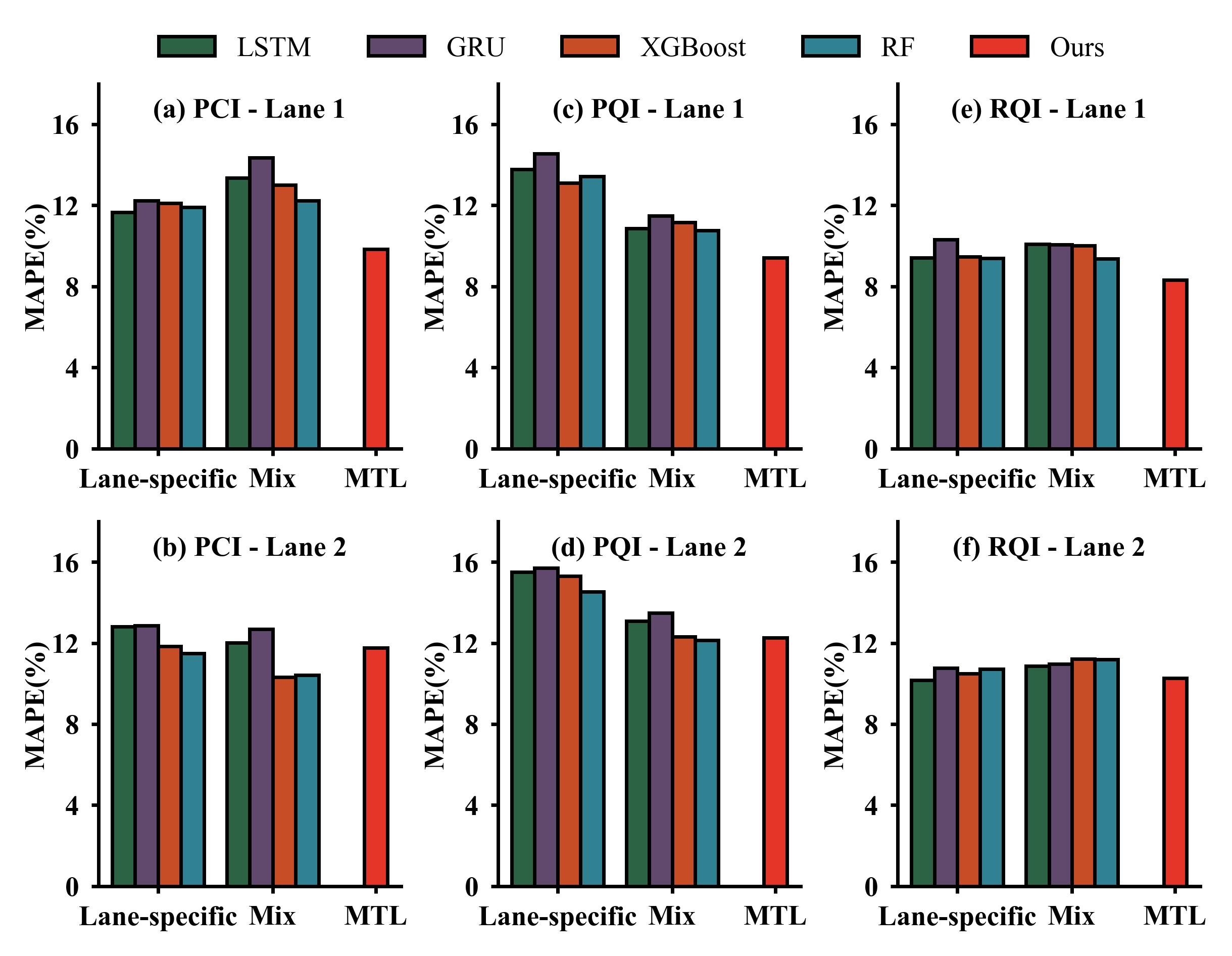}
\caption{The lane-level MAPE on the 2-lane datase}\label{fig5}
\end{figure}

\begin{figure}[htp]
\centering
\includegraphics[width=1\textwidth]{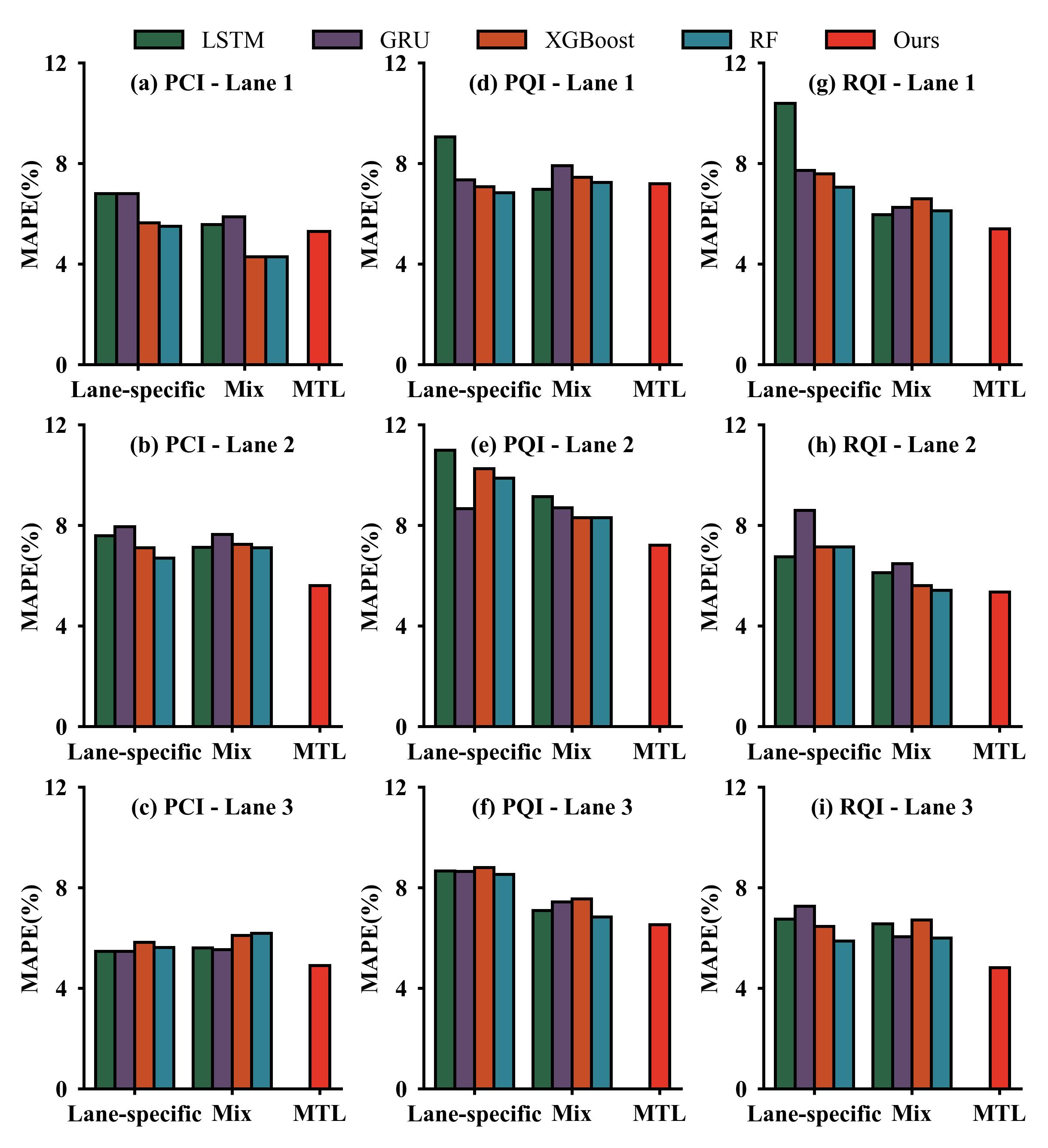}
\caption{The lane-level MAPE on the 3-lane dataset}\label{fig6}
\end{figure}

\begin{figure}[htp]
\centering
\includegraphics[width=0.99\linewidth]{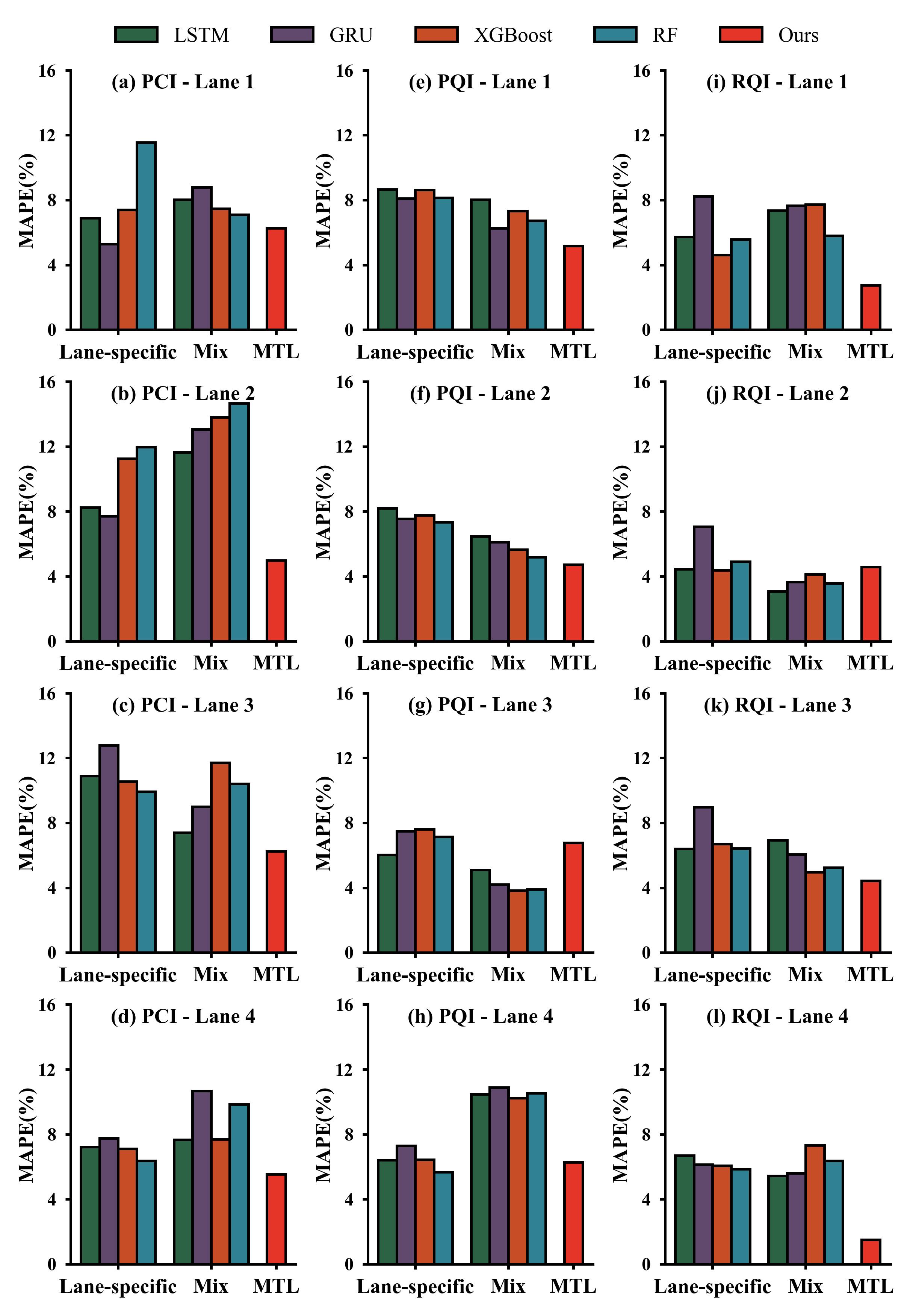}
\caption{The lane-level MAPE on the 4-lane dataset}\label{fig7}
\end{figure}

\subsection{Ablation experiment}
An ablation experiment is conducted on the model structure and pavement auxiliary features to assess the distinct contributions of each component on the 4-lane dataset. Regarding the model structure, three different test models were implemented. As for the auxiliary features, each feature was removed individually to assess its impact.

There are test models' introductions. In test model 1, the LSTM-based shared layer is removed, so the pavement performance time series are sent directly to each task-specific head. The shared layer typically extracts common pavement decay patterns from the input data, offering a universally applicable prediction outcome for all lanes. Furthermore, these layers incorporate a regularization effect that helps prevent over-fitting to any single task. In test model 2, the model eliminates the multiple task-specific heads. Instead, the output from the LSTM-based shared layer is transformed into a 2-dimensional format and then horizontally concatenated with auxiliary features. The presence of task-specific heads, each focusing on the prediction of a specific lane, can enhance prediction performance by concentrating on the lane-specific patterns most indicative of the outcomes for each lane. This specialization improves the accuracy and reliability of the predictions. In test model 3, the horizontal concatenation process is removed. The original model integrates pavement auxiliary features and initial predictions. The former enhances the model's ability to distinguish different prediction units. It is conducive to capturing the varied impacts on pavement performance deterioration more comprehensively, including factors such as traffic load and pavement material. The latter helps the model to adjust the predictions and get more precise results. It enables the model to detect more subtle and complex patterns. Moreover, feature concatenation provides increased regularization, which helps to mitigate the risk of over-fitting. 

The proposed model and test models are used to predict PCI, PQI, and RQI, and the MAPE results are shown in Figure 8(a). The results indicate that all the components contribute positively to enhancing the accuracy of the predictions. The LSTM-based shared layer is of the most significant importance to the proposed model. When this shared layer is removed, the MAPE increases by 1.631\%, 1.611\%, and 1.806\% when predicting PCI, PQI, and RQI, respectively. 

To validate the impact of every pavement auxiliary feature on the model's predictive performance, each auxiliary feature was removed individually, and the MAPE results are shown in Figure 8(b). The results indicate that all the utilized auxiliary features contribute positively to enhancing the accuracy of the predictions. However, there is not one feature that will greatly improve prediction accuracy. The reduction of MAPE ranges from 0.227\% to 0.547\% when predicting PCI, 0.300\% to 0.697\% when predicting PQI, and 0.245\% to 0.663\% when predicting RQI.

\begin{figure}[H]
\centering
\includegraphics[width=1\textwidth]{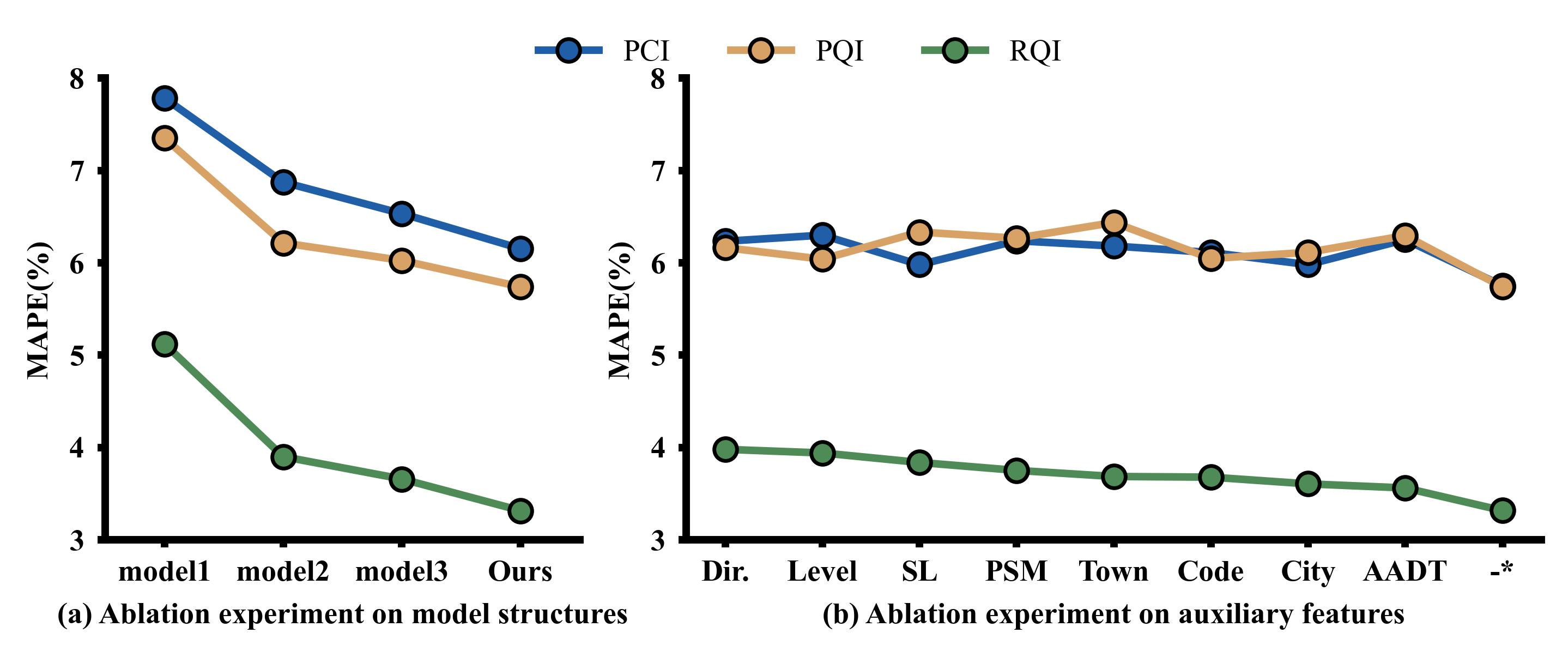}
\caption{The MAPE of ablation experiments}\label{fig8}
Note: - means there is no feature removed.
\end{figure}

\section{CONCLUSION}
The lane-level pavement performance prediction offers high-resolution information for developing maintenance strategies and plans. However,  there is a lack of high-resolution lane-level pavement performance data generally. Thus, this study proposes a multi-task deep learning approach that can predict pavement performance of multiple lanes in one unified model with enough historical segment-level data. The proposed deep learning approach mixed LSTM and ANN with a multi-task learning structure. Such design allows us to capture temporal evolution in pavement performance at both segment-level and lane-level, but also unify multiple prediction tasks into one modeling structure. Rigorous validation and ablation experiments have been conducted utilizing real pavement performance inspection data obtained from three cities within Henan Province, China: Zhengzhou, Jiaozuo, and Luoyang.

(1) The proposed model achieves an average of 10.309\%, 5.816\%, and 4.936\% MAPE when predicting PCI, PQI, and RQI on the 2-lane, 3-lane, and 4-lane datasets and gets an average absolute maximum reduction in MAPE of 2.774\%, 2.699\%, and 3.092\%, compared with the 'lane-specific model' and 'mix model'. Besides, the model gets the lowest MAPE of each lane in most cases. 

(2) Ablation experiments on model structure show the LSTM-based shared layer, multiple task-specific heads, and horizontal concatenation process play an important supporting role in reducing prediction error, and the reduction are 1.631\%, 1.611\%, and 1.806\% when predicting PCI, PQI, and RQI, respectively. Ablation experiments on pavement auxiliary features show all the utilized auxiliary features contribute positively to enhancing the accuracy of the predictions. However, there is not one feature that will greatly improve prediction accuracy. The max reduction of MAPE is 0.547\%, 0.697\%,  and 0.663\% when predicting PCI, PQI, and RQI.

(3) In most cases, the MAPE of each lane is the lowest using the proposed model compared with the lane-specific and mix models. The outcomes show the effectiveness of the proposed model in leveraging related tasks to enhance the overall and lane-specific forecasting accuracy in pavement performance analysis. As the number of lanes increases, the difference in MAPE among lanes narrows. However, it is worth noting that increasing the number of tasks does not necessarily increase the magnitude of MAPE reduction.

Future research may encompass in-depth discussions on two unresolved issues. First, the spatial relationship of the prediction units should be further explored during the prediction. Since the adjacent prediction units, or two units with similar environment, usually have similar pavement deterioration patterns. Second, few road units may experience very sharp drops in pavement performance, which violates most common deterioration patterns. This is challenging in precisely predicting such sharp drops with almost all models in the literature. 

\section*{Author contributions}
\textbf{Bo Wang}: Writing – original draft, Visualization, Validation, Software, Methodology, Investigation. \textbf{Wenbo Zhang}: Writing – review \& editing, Conceptualization, Methodology, Supervision, Resources, Project administration. \textbf{Yunpeng Li}: Methodology.

\section*{Funding sources}
This work was supported by the National Key Research and Development Program of China [grant number 2022YFB4300300]




\newpage
\bibliographystyle{unsrt}  
\biboptions{sort&compress}
\bibliography{elsarticle}
\end{document}